\documentclass[10pt,twocolumn,letterpaper]{article}

\usepackage{cvpr}
\usepackage{times}
\usepackage{epsfig}
\usepackage{graphicx}
\usepackage{amsmath}
\usepackage{amssymb}
\usepackage{multirow}
\usepackage{bbm}
\usepackage{mathrsfs}
\usepackage{booktabs}
\usepackage{subfigure}
\usepackage{xfrac}
\usepackage{esvect}

\usepackage[draft=False,pagebackref=true,breaklinks=true,letterpaper=true,colorlinks,bookmarks=false]{hyperref}

\cvprfinalcopy 

\def\cvprPaperID{4085} 

\ifcvprfinal\fi
\begin{document}

\title{StyleRemix: An Interpretable Representation for Neural Image Style Transfer }


\author{Hongmin Xu$^{*}$, 
Qiang Li$^*$, 
Wenbo Zhang, 
Wen Zheng\\
Y-tech, Kwai\\
}

\maketitle
\begin{abstract}
Multi-Style Transfer (MST) intents to capture the high-level visual vocabulary of different styles and expresses these vocabularies in a joint model to transfer each specific style. Recently, Style Embedding Learning (SEL) based methods represent each style with an explicit set of parameters to perform MST task. 
However, most existing SEL methods either learn explicit style representation with numerous independent parameters or learn a relatively black-box style representation, which makes them difficult to control the stylized results.~In this paper, we outline a novel MST model, StyleRemix, to compactly and explicitly integrate multiple styles into one network. 
 By decomposing diverse styles into the same basis, StyleRemix represents a specific style in a continuous vector space with 1-dimensional coefficients.
With the interpretable style representation,~StyleRemix not only enables the style visualization task but also allows several ways of remixing styles in the smooth style embedding space.~Extensive experiments demonstrate the effectiveness of StyleRemix on various MST tasks compared to state-of-the-art SEL approaches.
\footnote{*The first two authors contributed equally to this work.}
\end{abstract}

\section{Introduction}

Style transfer aims to convincingly confer artist's painting style, such as the shapes, lines, colors, tones, and textures created by the unique techniques of artist, to arbitrary images. With style transfer, people can produce Van Gogh's \emph{The Starry Night} style in their photographs. In the past, it may take much time of a well-trained artist to create such paintings. The critical issue of style transfer and its previous closely related work texture synthesis is how to model visual texture of target style image and generate a stylized image, which has similar visual fashion with the target style image while retaining the shape of input content image. Traditional methods \cite{efros1999texture, wei2000fast} usually focused on modeling texture with low-level features to handle this problem.


\begin{figure}
\begin{center}
\includegraphics[width= 1.0 \linewidth]{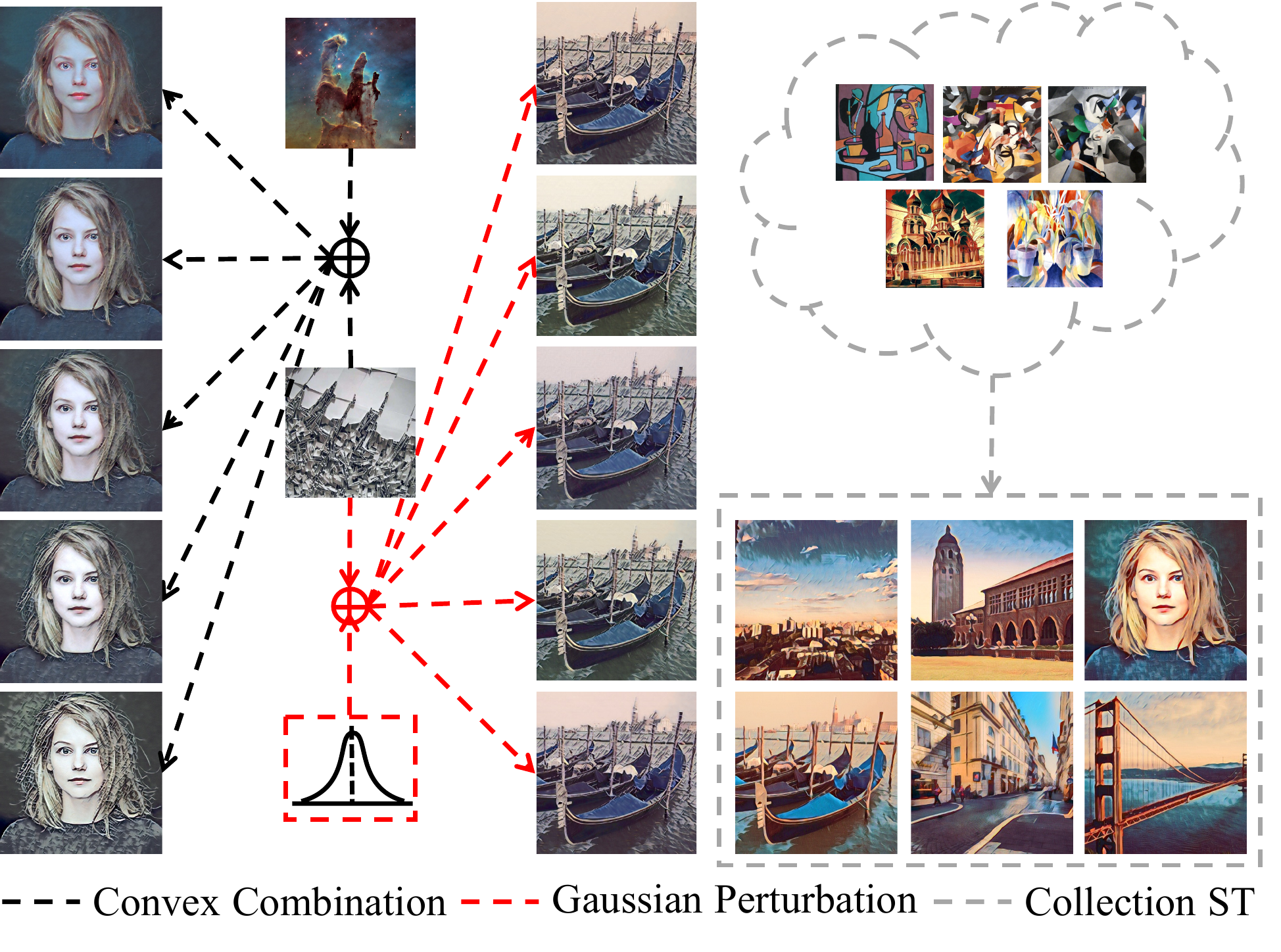}
\end{center}
   \caption{An example of StyleRemix in terms of Convex Combination (CC), Noise Perturbation (NP), and Collection Style Transfer (CST). CC offers a smooth transition between styles. NP offers a way to modify the style effect with noise. In the above example, the Gaussian noise is $\varepsilon \sim N(\mu, \sigma^2)$, where $\mu=\sfrac{1}{256}$ and $\sigma=0.005$. The style of CST is defined by multiple styles. }
\label{fig:SR_Title}
\end{figure}

The seminal work of Gatys \cite{gatys2015texture, gatys2016image}, Neural Style (NS) transfer, first modeled texture and style by the summary statistics in the domain of Convolutional Neural Network (CNN). Specifically, \emph{style} is defined by the covariance matrix of feature response computed by multi-layers of VGG-19 \cite{simonyan2014very}. 
Based on online powerful iterative optimization, NS outputs a stylized image, whose \emph{style} is close to the target image while retaining the shape of input image. Although NS produces better visually satisfied image than traditional methods, NS is time-consuming since its expensive computation cost. 
Therefore, Fast Neural Style (FNS) \cite{johnson2016perceptual, ulyanov2017improved, wang2017multimodal,ulyanov2016texture} transfer approaches trained offline feed-forward style-specific model to produce stylized images, which is hundreds of times faster than the NS method. 
However, FNS loses the flexibility of NS: different generative networks have to be trained for each specific style image. 

In order to improve the flexibility of FNS, Multi-Style Transfer (MST) methods aim to incorporate multiple styles into one single model. There are two paths to achieve this goal: 1) parametric Arbitrary Style Transfer (AST) approaches \cite{zhang2017multi, ghiasi2017exploring, shen2018neural, sheng2018avatar, li2018learning, huang2017arbitrary, wang2017zm, chen2016fast, gu2018arbitrary} and 2) non-parametric Style Embedding Learning (SEL) based style transfer approaches, such as \cite{chen2017stylebank, dumoulin2017learned, li2017diversified, yanai2017conditional}. 

AST methods \cite{zhang2017multi, sheng2018avatar, li2018learning, huang2017arbitrary, wang2017zm, chen2016fast, gu2018arbitrary} often learn a parametric network to describe a large set of styles and manually design \cite{chen2016fast, gu2018arbitrary, li2018learning, zhang2017multi, sheng2018avatar} or learn a network as a universal style transfer function \cite{wang2017zm, huang2017arbitrary, sheng2018avatar}. Therefore, the scalability of AST is larger than SEL to some extend. However, AST suffers from two shortcomings: 1) They always produce compromised visual quality stylized images, as some new styles may be poorly represented by the parametric style network or the designed/learned style transfer function may not be suitable for the new styles. 2) The style representation learned by a black-box network is relatively hard to explain and difficult to control the stylization.


%

In parallel, SEL methods \cite{chen2017stylebank, dumoulin2017learned, li2017diversified, yanai2017conditional} focused on learning a small set of parameters to explicitly define each style, which enable them to generate better visually impressive stylized results than AST. 
Dumoulin \etal \cite{dumoulin2017learned} utilized the parameters of instance normalization layers to represent one specific style. Li \etal \cite{li2017diversified} defined a noise map to indicate one particular style. Yanai \etal \cite{yanai2017conditional} used a simple one-hot map to denote different styles. However, those style representations are relatively hard to interpret.
As for Chen \etal \cite{chen2017stylebank}, they proposed an explicit style representation method, \emph{StyleBank}, which is more explainable 
 whereas \emph{StyleBank} need a layer of convolution filter banks to represent each style.  Therefore, for all the state of the art SEL methods, either they \cite{dumoulin2017learned, li2017diversified, yanai2017conditional} are hardly interpretable or  \cite{chen2017stylebank, yanai2017conditional} inefficient to represent styles.

In this paper, we propose a SEL method \emph{StyleRemix} that brings interpretability and efficiency to the style representation in MST task. Specifically, \emph{StyleRemix} separates \emph{content} (of input image) and \emph{style} (of target image) representation. 
\emph{Content} is transformed into multi-layers activations of an autoencoder. \emph{Style} is described as weighted convolution filters, which is convolved with the intermediate feature embedding (\emph{content}) to get the final stylized image. \emph{Style} representation consists of two parts: \emph{style basis} and \emph{style weights}.
	\emph{Style basis} correspond to interpretability and their channels describe the fundamental style elements. \emph{Style weights} correspond to efficiency and a specific style is modeled as a 1-dimensional continuous vector, which is more efficient than the block-wise representation method in \cite{chen2017stylebank}. 


Our method learns the style embedding space for a set of styles. Thus, MST task can be performed by providing the inference network with the style weights of a specific style. 
With the learned style representations,
\emph{StyleRemix} makes it possible for the task of style visualization.
Besides, thanks to the interpretability and efficiency of the style representation, our method offers several ways to manipulate styles, such as convex combination, noise perturbation, and even Collection Style Transfer (CST), as shown in Fig. \ref{fig:SR_Title}. Furthermore, StyleRemix can handle a certain amount of diverse styles with consistent visual quality according to a scalability analysis on the number of styles.






















	
	
	
	
	
	
	
	
	
%

\begin{figure*}
\begin{center}
\includegraphics[width= 1.0 \linewidth]{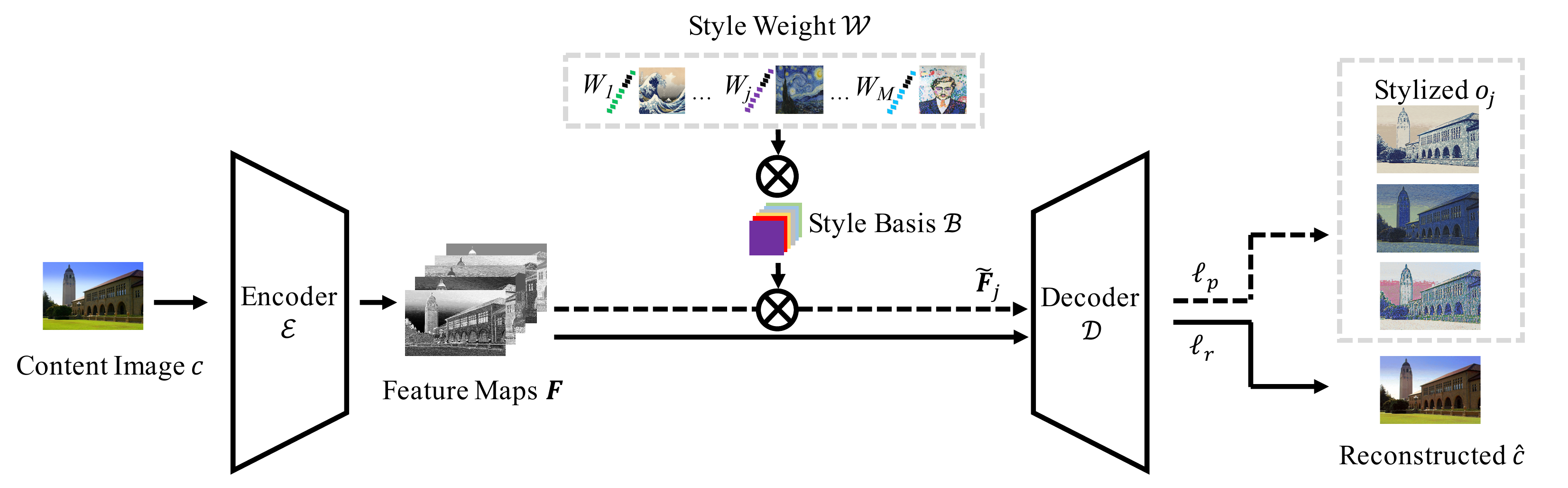}
\end{center}
   \caption{StyleRemix framework consists of an autoencoder (content image encoder $\mathcal{E}$, content image decoder $\mathcal{D}$), a style basis layer $\mathcal{B}$, and style weights layers $\mathcal{W}$. The weighted style basis can be achieved by a group-wise convolution operation.}
\label{fig:SR_Framework}
\end{figure*}

\section{Related Works}

\paragraph{Paired Image-to-Image Translation.} Image-to-Image Translation can be dated back to image analogy \cite{hertzmann2001image}. The goal of image analogy is modeling the transfer for a single paired input and output image. A recent work \cite{liao2017visual} investigated the image analogy problem in deep feature level. 
On the other hand,  the goal of image-to-image translation is learning the transfer function between the input image set and output image set.
For paired image-to-image translation, recent approaches used a paired dataset of input and output images to learn a parametric transfer function by using CNNs \cite{long2015fully}. Isola \etal \cite{isola2017image} used conditional generative adversarial network (GAN) \cite{goodfellow2014generative} to learn the transfer function between two sets.  With the similar idea, Sangkloy \etal \cite{sangkloy2017scribbler} generated images conditioned on both sketched boundaries and
sparse color strokes. 
Karacan \etal \cite{karacan2016learning} synthesized images from the semantic layout.
However, how to get the paired dataset is a tough problem in practice. 


\paragraph{Unpaired Image-to-Image Translation.} 
To handle the issue of obtaining paired dataset, unpaired image-to-image translation \cite{kim2017learning, liu2017unsupervised, zhu2017unpaired} methods have been proposed. The goal is tackling the unpaired data setting and learn the transfer function between two domains. To achieve this goal, Liu \etal \cite{liu2016coupled} shared the weights of the generators of different GANs to learn the joint representation in cross domains. Later, Liu \etal \cite{liu2017unsupervised} extended the framework of \cite{liu2016coupled} by combining the variational autoencoders \cite{kingma2013auto} with generative adversarial network.  
Another line of works, Zhu \etal \cite{zhu2017unpaired} and Kim \etal \cite{kim2017learning}, applied a cycle consistency loss to preserve the key attributes between two domains.



Concurrently, 
Collection Style Transfer (CST) is another approach to perform unpaired image-to-image translation task. The goal of CST is to compose the stylized image by taking advantage of multiple style images.
Sanakoyeu \etal \cite{sanakoyeu2018style} handled the CST task with the help of a style-aware loss and an adversarial loss to improve the stylization results.
The difference between MST and CST is that MST can transfer any element in the content image set into a specific style. However, the style of CST is defined by all style images in the style image set, and CST usually doesn't offer a method that can transfer the input content image into a particular style.


\section{StyleRemix Network}
The goal of style transfer can be defined as finding a stylized image $o$, which contains the visual fashion of target image $s$ while retaining the content of the input image $c$. For MST task, recent methods often use two sets: content image set $C=\{c_1, c_2, \ldots, c_N\}$  and style image set $S=\{s_1, s_2, \ldots, s_M\}$, to perform style transfer task. More specifically, MST methods aim to offer a model, which can transfer any element of the content set $C$ to a specific style $s_j$ in the style set $S$.


\subsection{Overview}
\emph{StyleRemix} aims to learn a compact and interpretable continuous vector representation for each style. The framework of \emph{StyleRemix} is shown in Fig. \ref{fig:SR_Framework}, which consists of three parts: an autoencoder, (\ie encoder $\mathcal{E}$ and decoder $\mathcal{D}$), a style basis layer $\mathcal{B}$, and style weights layers $\mathcal{W}$. 
In \emph{StyleRemix}, all styles in style set are decomposed and stored in a shared convolution layer and each particular style  is constructed by a weighted style basis.
As a result, we can simply represent each specific style with its style weights given the shared style basis. 
Finally, the weighted style basis is convolved with the intermediate feature maps \cite{hinton2006reducing} of input content image to obtain the stylized image $o$.
	
\subsection{Encoder and Decoder}  
We do not set any limitation on the architecture of encoder and decoder as shown in Fig. \ref{fig:SR_Framework}. Since the stylized process in StyleRemix is the weighted style basis directly operating on the intermediate feature maps of the input content image given by autoencoder. Any symmetry or asymmetry autoencoder architecture can be used to produce the intermediate feature maps $\boldsymbol{F} = \mathcal{E}(c)$. For simplicity, we use a 3 layers symmetry encoder and decoder, which is similar to \cite{johnson2016perceptual, chen2017stylebank}.

\subsection{Style Basis and Style Weights}  
\label{para:greetings}
	\paragraph{Style Basis.} By sharing the redundant convolution operators, which may be used to produce similar texture patterns, coarsening or softening strokes among different styles, style basis $\boldsymbol{B}$ is introduced to describe an embedding space of the style set. By doing so, a specific style $s_j$ in StyleRemix is described as a layer of weighted convolution filters, which map the high-level features of target style $s_j$ to input image $c$. The transferred feature maps $\widetilde{\boldsymbol{F}}_j$ given by style $s_j$ can be achieved by the convolution operation of the weighted style basis $\boldsymbol{B}_j$ over the intermediate feature maps $\boldsymbol{F}$:
	\begin{equation}
		\label{equ:stylized_mid_feature}
		\widetilde{\boldsymbol{F}}_j = \boldsymbol{B}_j \otimes \boldsymbol{F},
	\end{equation}
	where $\boldsymbol{F} \in \mathbb{R}^{c_{in} \times h \times w}$, $\boldsymbol{B}_j \in \mathbb{R}^{c_{out} \times c_{in} \times k_h \times k_w}$, $\widetilde{\boldsymbol{F}}_j \in  \mathbb{R}^{c_{out} \times h \times w}$, $c_{in}$ and $c_{out}$ are numbers of feature channels for $\boldsymbol{F}$ and $\widetilde{\boldsymbol{F}}_j$ respectively, $(h,w)$ is feature map size, and $(k_h, k_w)$ is the kernel size. The weighted style basis $\boldsymbol{B}_j$ is a point in learned style embedding space. In another word, $\boldsymbol{B}_j$ is determined by its corresponding coordinates, \ie style weights $\boldsymbol{w}_j$. Therefore, $\boldsymbol{B}_j$ is defined as:
	\begin{equation}
		\label{equ:weighted_style_basis}
		\boldsymbol{B}_j = (\mathbbm{1} \boldsymbol{w}_j^T) \odot \boldsymbol{B},
	\end{equation}
	where $\mathbbm{1} \in \mathbb{R}^{c_{out} \cdot k_h \cdot k_w \times 1}$, $\boldsymbol{w}_j \in \mathbb{R}^{c_{in} \times 1}$, and $\boldsymbol{B} \in \mathbb{R}^{c_{out} \cdot k_h \cdot k_w \times c_{in}}$, and $\odot$ is element-wise product. $\boldsymbol{B}_j$ in equation \ref{equ:stylized_mid_feature} and \ref{equ:weighted_style_basis} are equivalent by reshaping operation. 
	
	In StyleRemix, style basis $\boldsymbol{B}$ is implemented as a single convolution layer $\mathcal{B}$, which locates at the end of encoder $\mathcal{E}$ and the beginning of decoder $\mathcal{D}$ as shown in Fig. \ref{fig:SR_Framework}.

	\paragraph{Style Weights.} As we mentioned above, one purpose of style basis is compressing the redundancy of different styles. On the other hand, we hope that style basis itself should be interpretable. 
In another word, if the style base has no contribution to a specific style, its coefficient should be close to zero. To deliver this intuition, all weighted style basis are forced to be a convex combination of style basis $\boldsymbol{B}$. That is to say, style weights $\boldsymbol{w}_j$ lies in the simplex:
	\begin{equation}
		\begin{aligned}
		\label{equ:simplex_constraint}
		& \qquad \Delta_{c_{in}} = \{ \boldsymbol{w}_j \in \mathbb{R}^{c_{in} \times 1} \}, \\
		& \text{ s.t. } ~ \boldsymbol{w}_j[i] \ge 0 ~ \text{and} ~ \sum_i \boldsymbol{w}_j[i] = 1. \\
		\end{aligned}
	\end{equation}

	Style basis and style weights are jointly trained during one optimization iteration step. There are many ways to learn style weights while meeting the constraint in Eq. \ref{equ:simplex_constraint}. For simplicity, we implement style weights as a learnable layer $\mathcal{W}$, which consists of two sublayers: a linear layer without bias $\mathcal{L}$ and a softmax layer $\mathcal{S}$. The function of linear layer is defined as:
	\begin{equation}
		\label{equ:linear_layer}
		\boldsymbol{y} = a\boldsymbol{x},
	\end{equation}
	where $a$ is a scalar, $\boldsymbol{x}, \boldsymbol{y} \in \mathbb{R}^{c_{in} \times 1}$, 
	The function of softmax layer is defined as:
	\begin{equation}
		\label{equ:softmax_layer}
		y_i = \frac{\exp(y_i)}{\sum_j \exp(y_j)},
	\end{equation}	
	where $y_i$ is $i$th element in $\boldsymbol{y}$. By combining Eq. \ref{equ:linear_layer} and \ref{equ:softmax_layer} and setting $a=1$, style weights can be achieve by: $ \boldsymbol{w} = 1 \to \mathcal{L} \to \mathcal{S}$. It is worth noticing that  one of the state of the art MST framework, StyleBank\cite{chen2017stylebank}, can be regarded as a special case of StyleRemix. To be specific, in StyleBank, one can combine all style basis together, and set the style weights to a block-wise one-hot vector which is not learnable.
	
	Since the style $s_j$ is modeled by the coordinates in the learned style embedding space, style manipulation can be achieved by simply operating style weights. For instance, the convex combination among styles can be performed by the convex combination of the style weights. Besides, it is possible to produce various stylized results for a particular style $s_j$ by adding noise to its corresponding style weights. Furthermore, since style basis are learned to represent the styles of the style set, \emph{StyleRemix} even allows to achieve CST task by, such as, simply averaging the style weights vectors or offering the same value to all style weights. In that way, the final stylized results benefit from all style basis, which is the goal of CST.
	
	\begin{figure}[t]
		\begin{center}
   		\includegraphics[width=1.0\linewidth]{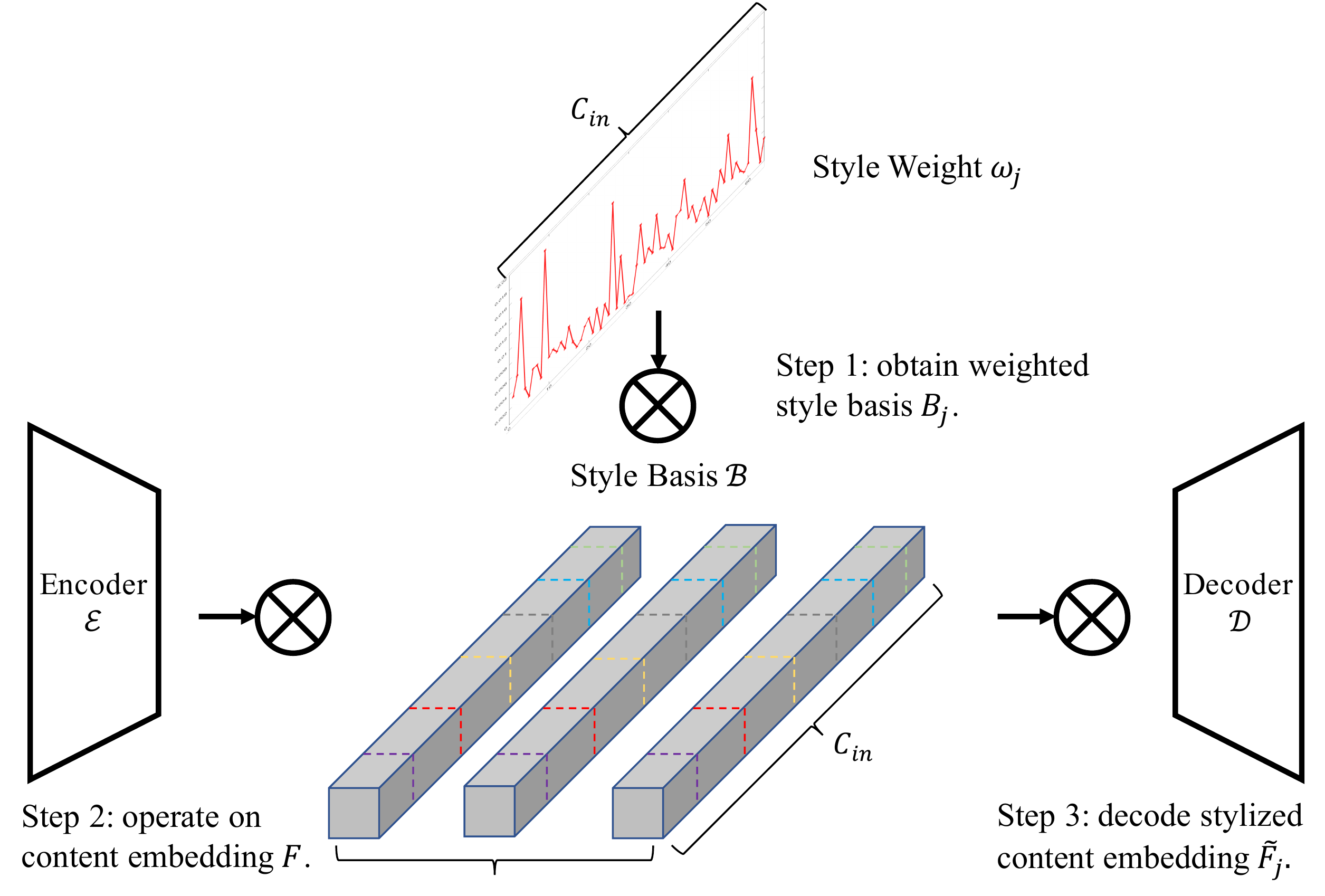}
		\end{center}
		\caption{An illustration of the forward method of the stylizing branch which includes three steps.}
		\label{fig:SR_Forword_method}
	\end{figure}

\subsection{Loss Function.} StyleRemix includes two training branches: autoencoder branch (\ie $\mathcal{E} \to \mathcal{D}$) and stylizing branch (\ie $\mathcal{E} \to (\mathcal{W} \to \mathcal{B}) \to \mathcal{D}$, where $\mathcal{W}: 1 \to \mathcal{L} \to \mathcal{S}$). In order to clearly understand the forward method of the stylizing branch, we show the example of how it works in Fig. \ref{fig:SR_Forword_method}. 

	
	For the autoencoder training branch, the \emph{Euclidean Distance} between the input content image $c$ and the reconstructed content image $\hat{c}$ is adopted to measure the reconstruction error, which can be achieved by:
	\begin{equation}
		\label{equ:pixel_loss}
		\mathcal{L}_r(\hat{c}, c) = ||\hat{c} - c ||^2_2.
	\end{equation}
	
	For stylizing training branch, the $perceptual$ $loss$ defined in \cite{johnson2016perceptual} is used to learn style representation. The $perceptual$ $loss$ can be achieved by:
	\begin{equation}
		\label{equ:perceptual_loss}
		\mathcal{L}_p(\hat{c}, c, s_j) = \alpha \mathcal{L}_c(\hat{c}, c)  + \beta \mathcal{L}_s(\hat{c}, s_j),
	\end{equation}
	where $s_j$ is a specific style belonging to the style set $S$. 
	
	Content loss $\mathcal{L}_c$ and style loss $\mathcal{L}_s$ are respectively defined to measure the difference of feature maps and gram matrix \cite{gatys2016image} between different images. Both feature maps and gram matrix are computed from VGG-16 network \cite{simonyan2014very}. Content loss $\mathcal{L}_c$ and style loss $\mathcal{L}_s$ can be respectively achieved by:
	\begin{align}
		\label{equa:content_style_loss}
		& \begin{matrix} \mathcal{L}_c(\hat{c}, c) \end{matrix} = \sum_{l \in \{l_c\}}||F^l(\hat{c}) - F^l(c) ||^2_2, \\
		& \begin{matrix} \mathcal{L}_s(\hat{c}, s_j) \end{matrix} = \sum_{l \in \{ l_s\} }\frac{1}{n_l}||G^l(\hat{c}) - G^l(s_j) ||^2_2,
	\end{align}
	where $F^l$ and $G^l$ are feature maps and gram matrix calculated from $l$th layer of VGG-16 network \cite{simonyan2014very}, $n_l$ is the total number of units of layer $l$, $\{l_c\}$ and $\{l_s\}$ are layers which deliberately chosen to obtain $F^l$ and $G^l$.

	\begin{figure}
		\centering
		\subfigure[Warming-up phase.]{
			\begin{minipage}[b]{0.45 \textwidth}
			\includegraphics[width=1\textwidth]{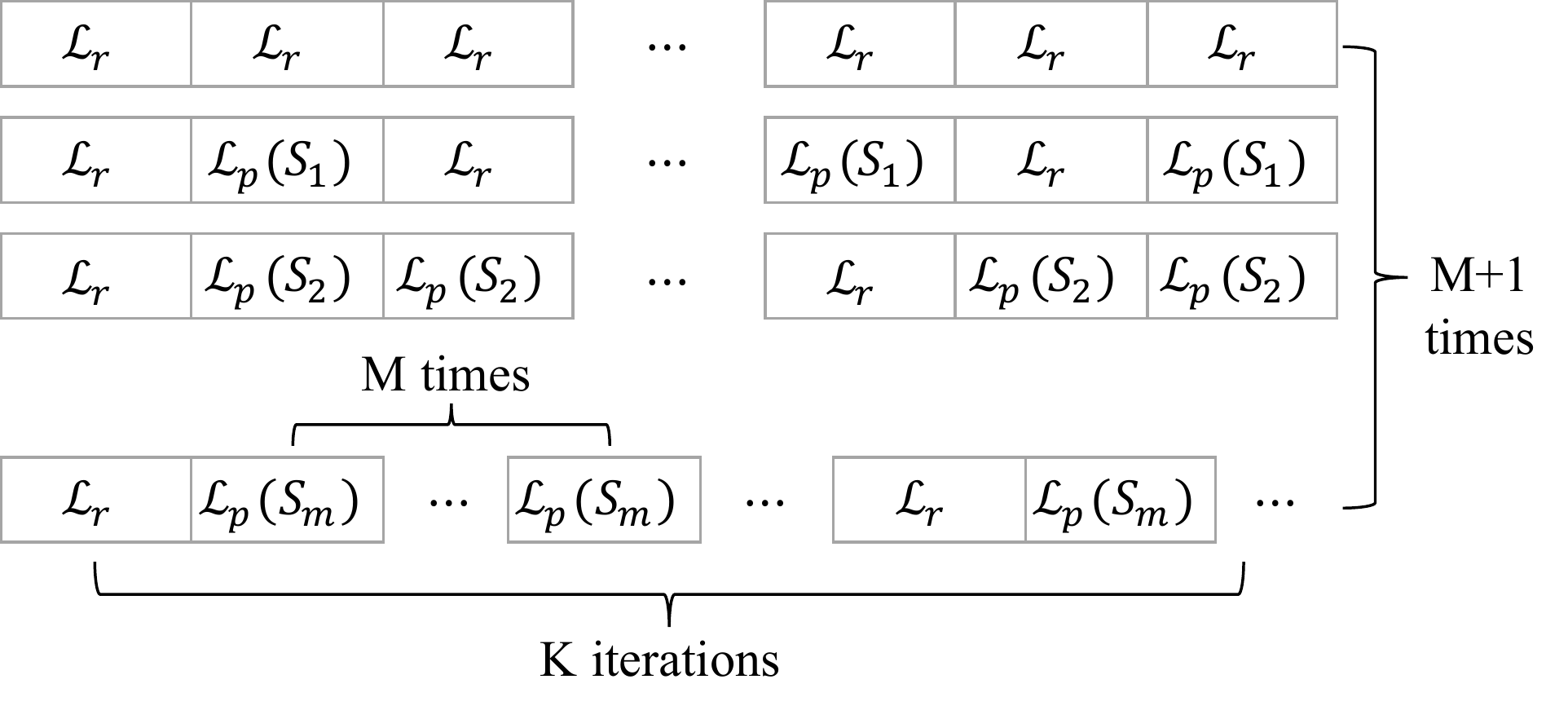}
			\end{minipage}
		}
		\subfigure[Finetuning phase.]{
			\begin{minipage}[b]{0.37 \textwidth}
			\includegraphics[width=1\textwidth]{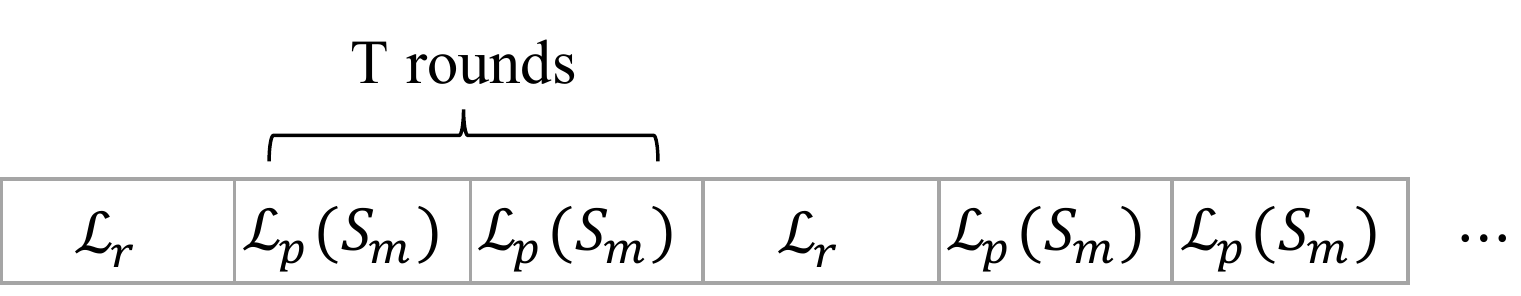}
			\end{minipage}
		}
	 	\caption{The detailed procedure of the warming-up and finetuning phases.}
		\label{fig:training_strategy_ISB_SL}
	\end{figure}

\subsection{Training Strategy.}

StyleRemix is difficult to train because the model has to convincingly express all the styles in the style set while comprehensively describing the content information.
Inspired by \cite{chen2017stylebank} and \cite{li2017diversified}, our training strategy includes two phases: a warming-up stage and a finetuning stage. 

The detailed training steps of the warming-up and the finetuning stages are shown in Fig. \ref{fig:training_strategy_ISB_SL}. 
The warming-up stage is trying to make the network get a general picture of the content and style information to be learned gradually.
For this purpose, in the warming-up stage, each new style will be incrementally added to the current style set, \ie from $S_1 = \{s_1\}, S_2 = \{s_1, s_2\}$ to $S_m = \{s_1, s_2, \ldots, s_m\}$. 
In the first $K$ iterations, only the autoencoder is trained. 
During the following each $K$ iterations, a new style set $S_j$ will be trained until no more new style is available. 

After the warming-up stage, we conduct the finetuning phase to enforce the network to strengthen the learned content and style characteristics. In particular, we employ the $T+1$ training strategy, namely $1$ round training on content images and T rounds training on the style set $S_m$ to capture the content and style characteristics simultaneously.

\begin{figure*}
\begin{center}
\includegraphics[width= 1.0 \linewidth]{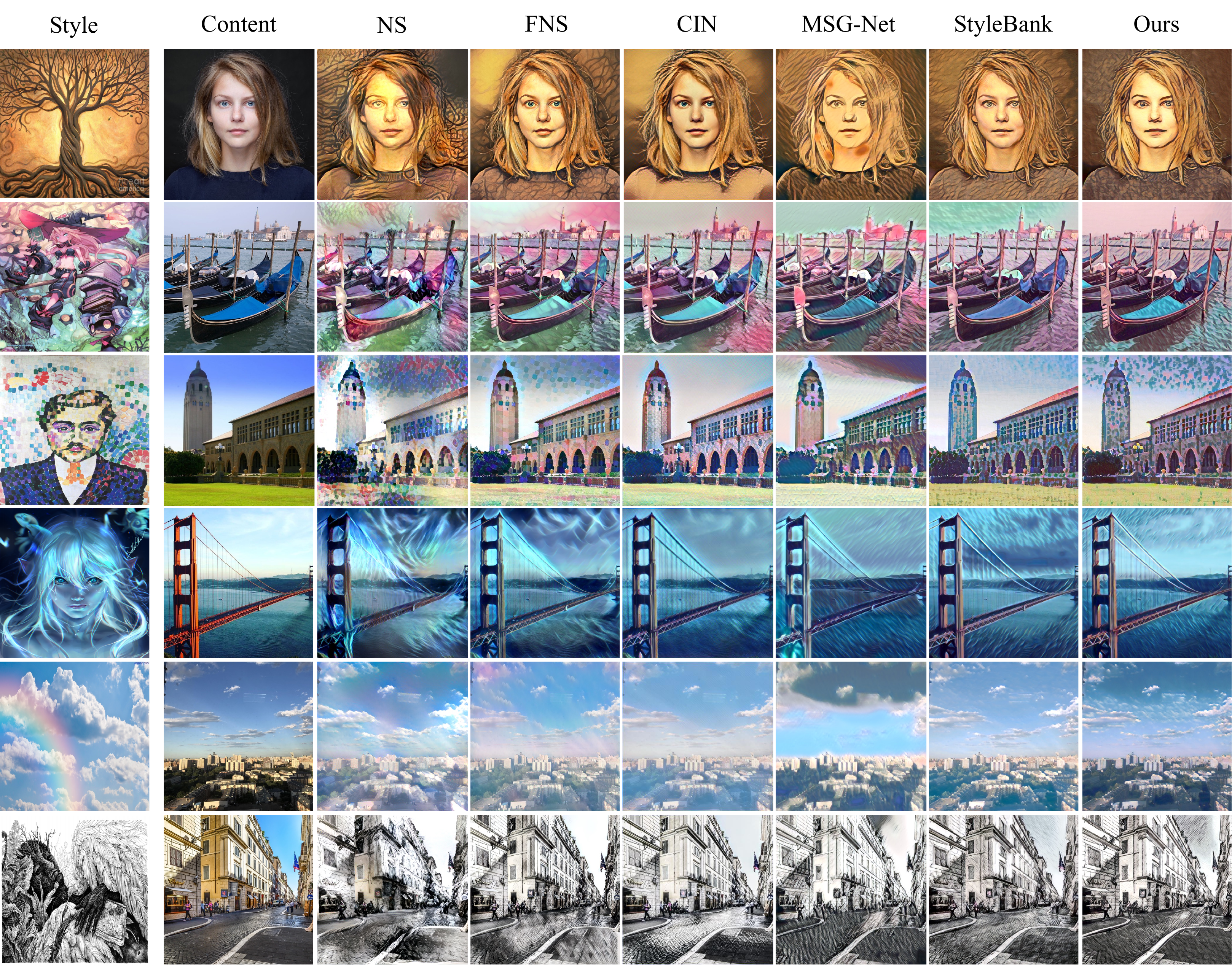}
\end{center}
   \caption{Some example results for qualitative evaluation.}
\label{fig:MST_comparison}
\end{figure*}

\section{Experiment}
\label{sec:experiment}

\paragraph{Implementation details.} 
We employ a simple symmetry autoencoder same to \cite{chen2017stylebank} but with 256 channels style basis. 
The detailed architecture of network is shown in Table \ref{tab:SR_Architecture}. The linear layer has 1 input channel and 256 output channels and the softmax layer has 256 input channels and 256 output channels.
We employed VGG-16 \cite{simonyan2014very}, pre-trained on the ImageNet dataset \cite{russakovsky2015imagenet}, as perceptual feature extractor. We respectively chose layer \emph{Conv2\_2} and layers \emph{Conv1\_2}, \emph{Conv2\_2}, \emph{Conv3\_2}, and \emph{Conv4\_2} for content and style losses. The parameters $\alpha$ and $\beta$ in Eq. \ref{equ:perceptual_loss} are respectively set to: $\alpha = 1$ and $\beta = 3e4$ for all results of \emph{StyleRemix}  in MST task except for specific statements. We randomly sample 1000 images from Microsoft COCO dataset \cite{lin2014microsoft} as content dataset. For style dataset, $S_{50}$ contains 50 style images which are widely used in other MST papers. 
 Note that another style dataset, which contains 50 paintings of \emph{Van Gogh}, is employed for CST task.

During the training process, content images are randomly cropped to $512 \times 512$, style images are scaled to 600 along large side. StyleRemix is trained with batch size  4, style weights initialized with the same value, parameters $T$ in style learning stage is set to 2, learning rate $0.001$ with decayed 0.8 at every $30K$ iterations, and Adam optimizer \cite{kingma2014adam} for $300K$ iterations. Our implementation is based on PyTorch \cite{paszke2017automatic}. 

\begin{table}[t]
	\centering
	\begin{center}
	\setlength{\tabcolsep}{1mm}{
	\begin{tabular}{ccc}
		\toprule
		Layer Type 						&	Activation Dimensions						\\
		\midrule
		Input								&	$H \times W \times 3$						\\
		$\text{Conv-IN-ReLU-S}_1$    			&	$H \times W \times 32$						\\
		$\text{Conv-IN-ReLU-S}_2$   			&	$\sfrac{1}{2}H \times \sfrac{1}{2}W \times 64$		\\
		$\text{Conv-IN-ReLU-S}_2$    			&	$\sfrac{1}{4}H \times \sfrac{1}{4}W \times 256$		\\
		$\text{Conv-IN-ReLU-S}_1$ (StyleBasis)   &	$\sfrac{1}{4}H \times \sfrac{1}{4}W \times 256$		\\
		$\text{Conv-IN-ReLU-S}_{\sfrac{1}{2}}$	&	$\sfrac{1}{2}H \times \sfrac{1}{2}W \times 64$		\\
		$\text{Conv-IN-ReLU-S}_{\sfrac{1}{2}}$	&	$H \times W \times 32$						\\
		$\text{Conv-S}_1$							&	$H \times W \times 3$						\\
		\bottomrule
	\end{tabular}
	}
	\end{center}	
	\caption{The symmetry architecture used in StyleRemix. Conv: Convolution layer. IN: Instance Normalization layer. ReLU: ReLU layer. $S_{n}: \text{stride} = n$.}
	\label{tab:SR_Architecture}
\end{table}


\subsection{Multi-Style Transfer}
We compare StyleRemix against several baselines to measure the performance of MST task. Baselines include: Neural style (NS) \cite{gatys2016image}, Fast Neural Style (FNS) \cite{johnson2016perceptual}, StyleBank \cite{chen2017stylebank}, Multi-Style Generative Network (MST-Net) \cite{zhang2017multi}, and Conditional Instance normalization (CIN) \cite{dumoulin2017learned}. StyleBank is trained using the same network and parameters as StyleRemix. Other baselines are trained with default parameters from their open source projects, except that NS is optimized for 4000 iterations, MSG-Net is trained for 5 epochs. Besides, all MST baselines are trained with the same style set $S_{50}$ as StyleRemix.

\label{sec:MST}

\paragraph{Qualitative Result.} 
The stylized images ($512 \times 512$) are shown in Fig. \ref{fig:MST_comparison}.
In general, our method has achieved competitive stylization results compared to other baselines.

More specifically, for Single-Style Transfer (SST) methods, since plenty of parameters can be used to describe a single style, SST methods often yield better results than other baselines. However, NS produces many artifacts result in the details of stylized results looking unnatural, \eg rows 1, 2, and 4. 
It may be caused by that the optimization process is based on the representation of the whole image.  FNS yields larger texture pattern, which even larger than textures in target style image, \eg rows 2, 3, and 4. 
This issue is possibly due to the reason that FNS uses $9\times9$ kernels and deep architecture which has a larger receptive field.

For MST methods, since the diversity of style set has a significant influence on the stylized results, they have to compromise with different styles. For CIN, it can be observed the color histogram shift only, and some weird textures scattered inconsistently throughout, \eg rows 1 and 3. Besides, CIN fails to reproduce texture patterns in style images. 
 One possible reason is CIN's style representation come from all the instance normalization layers, which do not explicitly separate content and style. For the same reason, CIN may be incapable of explicitly describing the textures of style.
For MSG-Net, it seems to transfer the same structure but different color textures, \eg rows 1, 2, and 3, and contain some large artifacts, \eg rows 1, 3, and 5. The root cause may be that MST-Net explicitly model the universal transfer function by a set of learnable weights. Likewise, MSG-Net suffers from some interferences among different styles.

Comparing with SST methods, StyleRemix produces appealing stylized results in a different way. To be specific, StyleRemix creates the fundamental elements of style, such as texture patterns, softening strokes, instead of the abstract color area produced by NS or FNS, \eg rows 2 and 5.
Besides, thanks to the separated representations of content and style, StyleRemix yields more visually pleasing results
than CIN and MSG-Net. For example, StyleRemix has captured textures in style images, \eg rows 1, 3, and 6. Since StyleBank is a special case of StyleRemix, their stylized results are visually similar.    

\begin{figure*}
	\begin{minipage}[t]{0.48\textwidth}
		\centering
		\includegraphics[width= \textwidth]{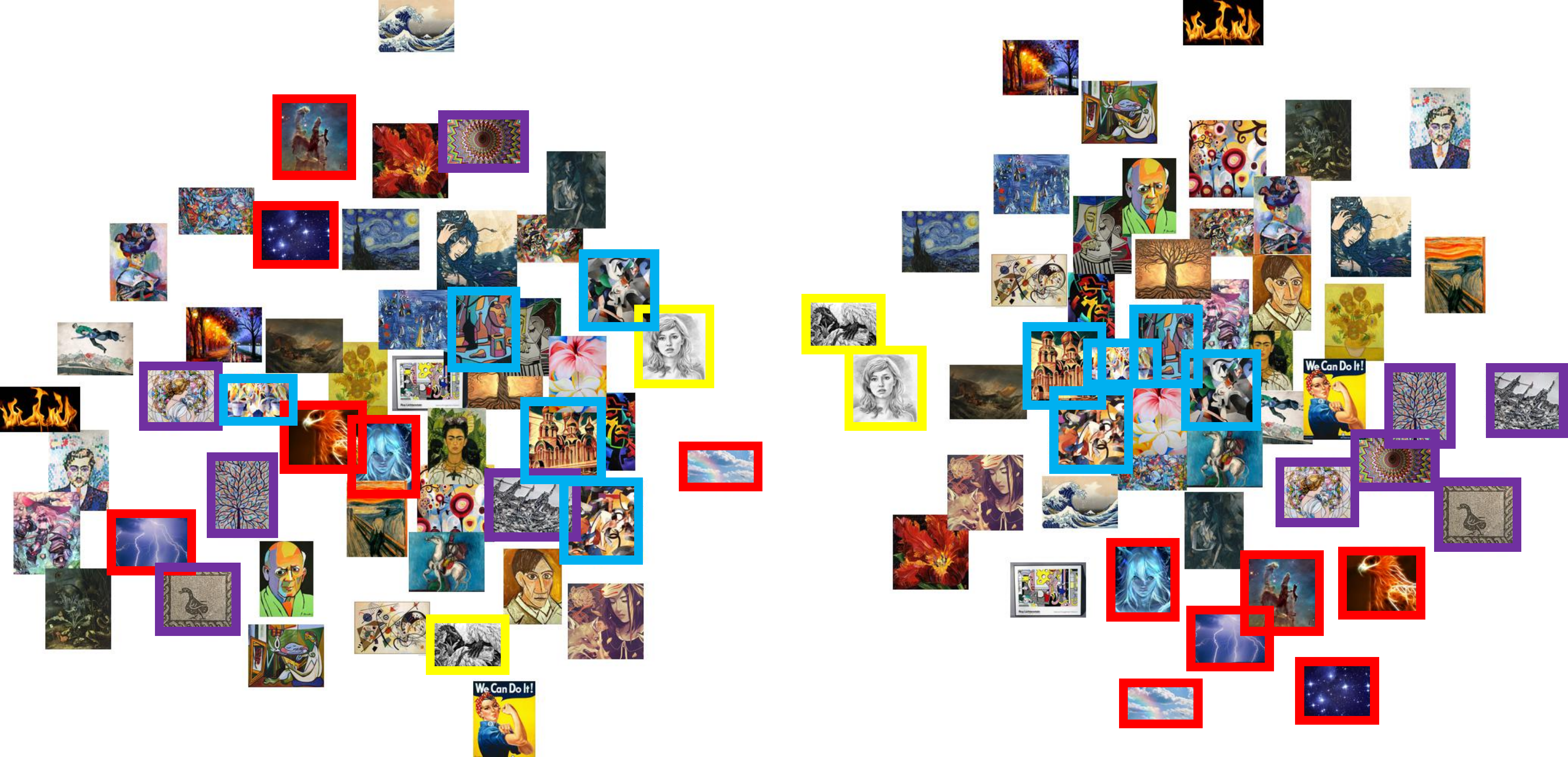}
		\caption{Structure of a 2 dimensional representation of the embedding space for CIN (\emph{left}) and StyleRemix (\emph{right}). 
		}
		\label{fig:style_t_SNE}
	\end{minipage}
	\hfill
	\begin{minipage}[t]{0.48\textwidth}
	\centering
		\includegraphics[width=\textwidth]{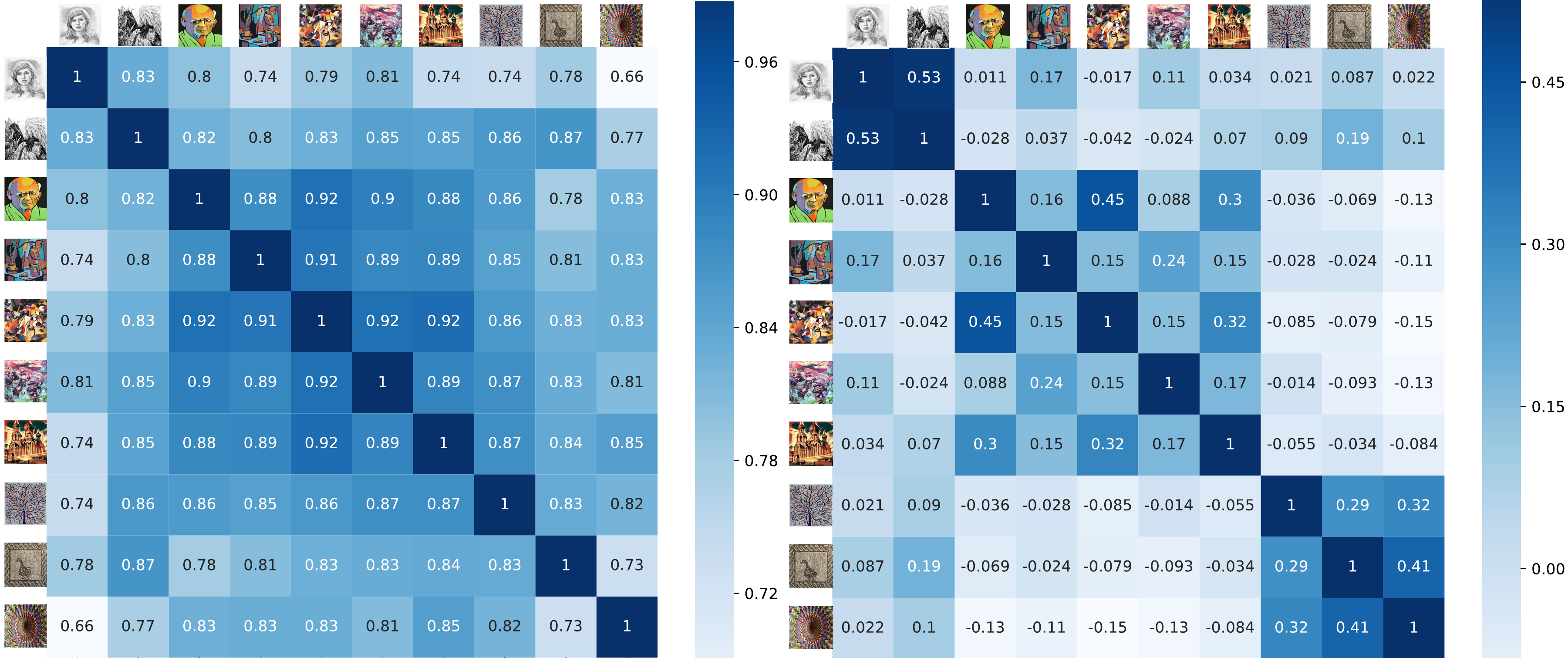}
		\caption{The correlation matrix of 10 chosen styles for for CIN (\emph{left}) and StyleRemix (\emph{right}).}
		\label{fig:style_corr}
	\end{minipage}%
\end{figure*}


\paragraph{Quantitative Results.} 
We compare the quantitative results of StyleRemix with other baselines in three aspects: Deception Rate \cite{sanakoyeu2018style} (DR), the number of float-point operations (FLOPs), and the cost of adding a new style (Cost/Style). 
The DR is calculated as the fraction of generated images which were classified by a network as the artworks of an artist for which the stylization was produced. The network is pre-trained on Wikiart to classify 624 artists.
For DR, we use the available 14 artists from the open source project of  Sanakoyeu \etal \cite{sanakoyeu2018style}. Since DR test mainly focus on paintings, we set the $\beta = 6e4$ for paintings stylization. And we generate 4200 stylized images (14 styles, 300 per style) for every method. 
FLOPs is calculated by counting the number of all convolution operations for single style. Since MSG-Net is based on the architecture of FNS \cite{johnson2016perceptual} and need an extra parametric network to obtain style representation, we didn't list its FLOPs. 
Cost/Style is defined by the parameters cost of adding a new particular style.
Results in Tab. \ref{tab:Quantitative_results} show that our method has competitive performance compared to other baselines.
Worth mentioning that, StyleRemix requires the lightest cost when adding a new style.



\begin{table}[t]
	\centering
	\begin{center}
	\setlength{\tabcolsep}{1mm}{
	\begin{tabular}{cccc}
		\toprule
		Method	 						&	DR (mean)		&	FLOPs				&	Cost/Style\\
		\midrule
		NS								&\textbf{0.2369}		&		-					&		-\\
		FNS								&	0.0532			&	$40,315MB	 $			&		-\\
		CIN								& 	0.0144			&	$40,315MB 	$			&		3206\\
		MSG-Net							&	0.0715			&		-					&		-\\
		StyleBank							&	0.1065			&	28,236MB					& $256^2\times3^2$ \\
		\textbf{StyleRemix}					&\textbf{0.1121}		&\textbf{28,236MB}				&	\textbf{256}	\\
		\bottomrule
	\end{tabular}
	}
	\end{center}
	\caption{Quantitative results comparison.}
	\label{tab:Quantitative_results}
\end{table}

\subsection{Style Visualization} 
According to our best knowledge, only CIN \cite{dumoulin2017learned} claims that they learn the embedding of style and more experiment results are presented in \cite{ghiasi2017exploring} to support this opinion. We compare the style embedding learned by StyleRemix and CIN. The style representation of StyleRemix is $\vv{S}_{sr} \in \mathbb{R}^{256 \times 1}$, and the style representation of CIN is $\vv{S}_{cin} = \{ \gamma_s, \beta_s\}, ~ \vv{S}_{cin} \in \mathbb{R}^{3206 \times 1}$, where $\gamma_s$ and $\beta_s$ are respectively the learned mean and standard deviation of instance normalization layers. 

We examine the low dimensional style representations of both CIN and StyleRemix. 
The style parameter of StyleRemix, in Eq. \ref{equ:perceptual_loss}, is set to $\beta = 6e4$,  to capture more information about style. Also, in order to achieve low dimensional display, we first employ Principal Component Analysis to reduce both style representation to 10 dimensions and then t-SNE \cite{maaten2008visualizing} dimensional reduction technique to further reduce the 10 dimensions style representation to 2 dimensions. Notice t-SNE will necessarily distort the representation significantly in order to compress the representation to small dimensionality. Therefore, we restrict our analysis to the qualitative description. The t-SNE results are shown in Fig. \ref{fig:style_t_SNE}. 

Comparing the t-SNE results of  CIN in Fig. \ref{fig:style_t_SNE} (left) and StyleRemix in Fig. \ref{fig:style_t_SNE} (right), one may notice StyleRemix has better locality than CIN. As shown in Fig. \ref{fig:style_t_SNE}, we use different color boxes to identify different class styles. For instance, for the t-SNE results of StyleRemix, in the yellow box, the styles contain only black, white color and both of them look like \textit{Sketch}. 

We also show the correlation matrix of 10 selected styles, which is directly calculated by original high dimensional style weights, in Fig.\ref{fig:style_corr}. The correlation matrix provides strong evidence that the style embedding learned by StyleRemix can distinguish styles in semantic similarity. The visually similar styles have high correlation coefficients, and the visually different styles have low correlation coefficients. Therefore, the correlation matrix of StyleRemix (Fig.\ref{fig:style_corr} right) shows \emph{Blocking Effect}, which is not observed in the correlation matrix of CIN (Fig. \ref{fig:style_corr} left).

\subsection{Style Manipulation} 
We conduct the \emph{Style Remixing} manipulation by decomposing all styles into the same style embedding space and recombining them to create a large number of new style effects. 
We show some heuristic examples to show how to remix styles via simply operating the style weights.

\begin{figure}
\begin{center}
\includegraphics[width= 1.0 \linewidth]{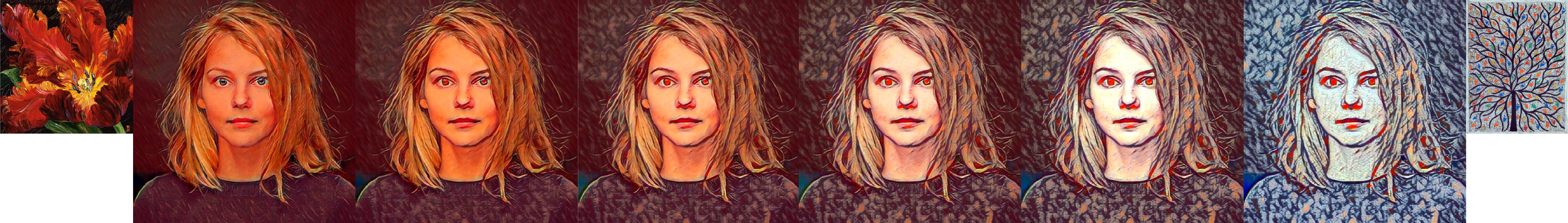}
\end{center}
   \caption{The convex combination results of two different styles. $\alpha = 1.0, 0.8, 0.6, 0.4, 0.2, 0.0$ respectively. }
\label{fig:convex_combination}
\end{figure}

\paragraph{Convex Combination.} Convex combination offers a smooth transition from one style to the other, which can be achieved by simply apply the convex combination in style weights of different styles. Supposing there are two different styles $s_l$ and $s_k$ with style weights $\boldsymbol{w}_l$ and $\boldsymbol{w}_k$, the convex combination style $s_{new}$ can be achieved by:
	\begin{equation}
		\boldsymbol{w}_{new} = \alpha \boldsymbol{w}_l + (1-\alpha) \boldsymbol{w}_k,
	\end{equation}
	where $0 \le \alpha \le 1$, $\boldsymbol{w}_{new}$ is the style weights of  $s_{new}$. Fig. \ref{fig:convex_combination} presents the convex combination results, of which one can observe the smooth transition between the two styles.

\begin{figure}
\begin{center}
\includegraphics[width= 1.0 \linewidth]{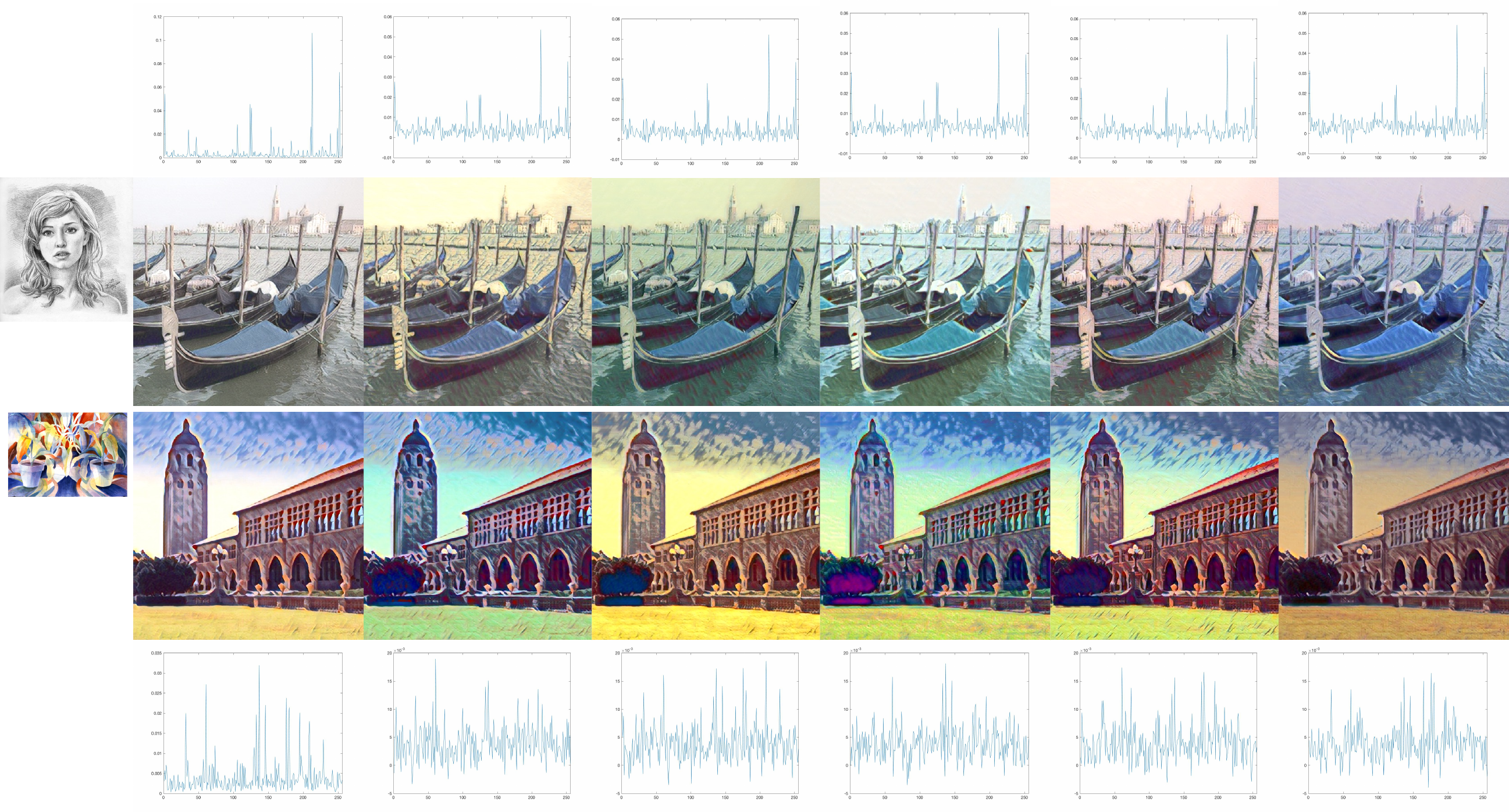}
\end{center}
   \caption{The perturbation study results of two different styles. The first column are style images. The second column are the original stylized results and its corresponding style weights. The following columns are obtained by adding Gaussian noise.}
\label{fig:Frequency_Selection} 
\end{figure}

\paragraph{Perturbation Study.} 
We show an example how to remix styles by adding Gaussian noise to a particular style. The results are shown in Fig. \ref{fig:Frequency_Selection}. We add Gaussian noise $\varepsilon \sim N(\mu, \sigma^2)$, where $\mu=\sfrac{1}{256}, \sigma = 0.005$, to the style weights $\boldsymbol{w}_j$. Then normalized the vector $\boldsymbol{v} = \boldsymbol{w}_{j} + \varepsilon$ by $\boldsymbol{v} = \boldsymbol{v} /  \sum^{256}_{i = 1} v_i$.
	One may notice that parts of the style weights of perturbed style are negative. Interestingly, it is difficult to differentiate the perturbed stylized results from the original results. These results may suggest that StyleRemix has captured a smooth embedding space, which is robust to noise to some extend.

\begin{figure}
\begin{center}
\includegraphics[width= 1.0 \linewidth]{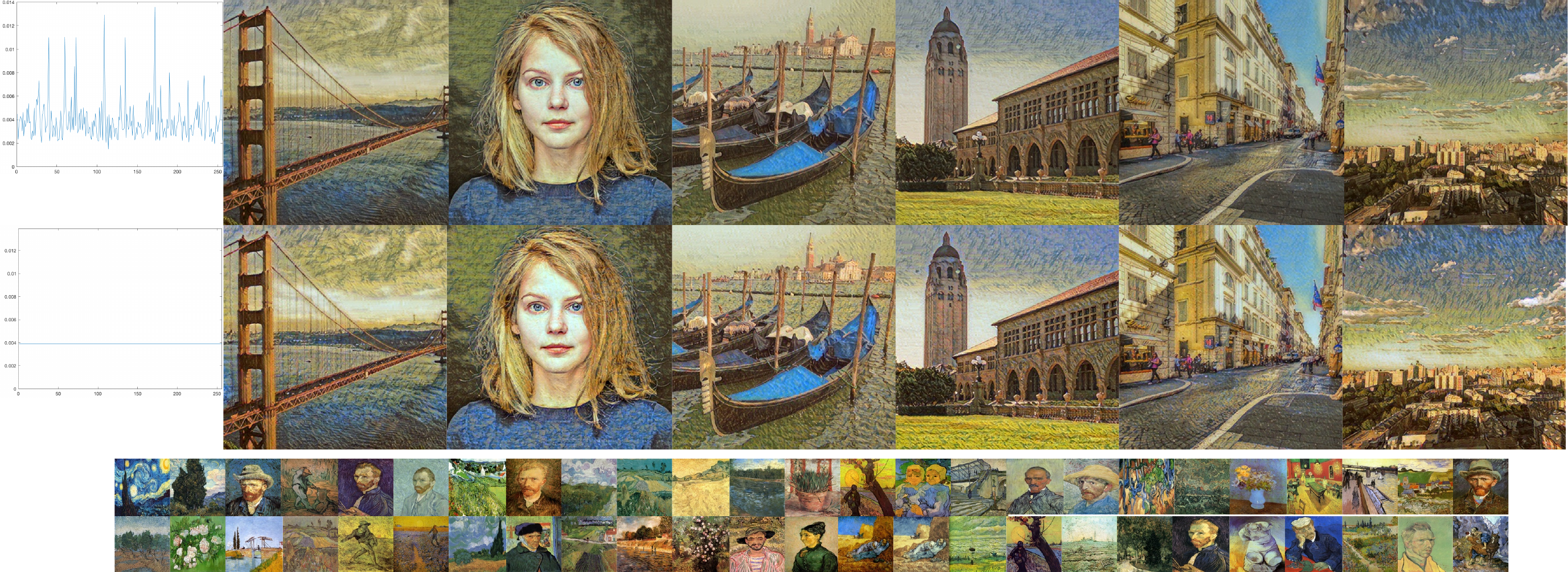}
\end{center}
   \caption{The CST results achieved by offering style weights. All 50 styles of \emph{Vincent Van Gogh} are shown in the last row.}
\label{fig:CST_Result} 
\end{figure}

\paragraph{Collection Style Transfer.} StyleRemix may offer a way to perform Collection Style Transfer (CST) task. As we mentioned before, the style of CST is defined by all styles in the style set. As the style elements of the style set are fully represented by the style basis in StyleRemix, it is possible to define the style of CST by manipulating the style weights. We train StyleRemix with 50 random paintings of \emph{Vincent Van Gogh} from Wikiart \cite{karayev2013recognizing}. The parameters $\alpha$ and $\beta$ in Eq. \ref{equ:perceptual_loss} are respectively set to: $\alpha = 1$ and $\beta = 6e4$. We show CST results in Fig. \ref{fig:CST_Result} with two simple manipulating strategies. One is to use the average vector of all 50 styles (row 1) as style weights, another is to set the same value $\sfrac{1}{256}$ (row 2) for all weights. They yield similar results as both are capturing all the style elements of Van Gogh paintings. On the other hand, they have subtle difference on chrominance, luminance and local texture as the style weights varied.

\subsection{Style Scalability}
\label{sec:StyleRemix_Capabilities} 
\begin{figure}
\begin{center}
\includegraphics[width= 1.0 \linewidth]{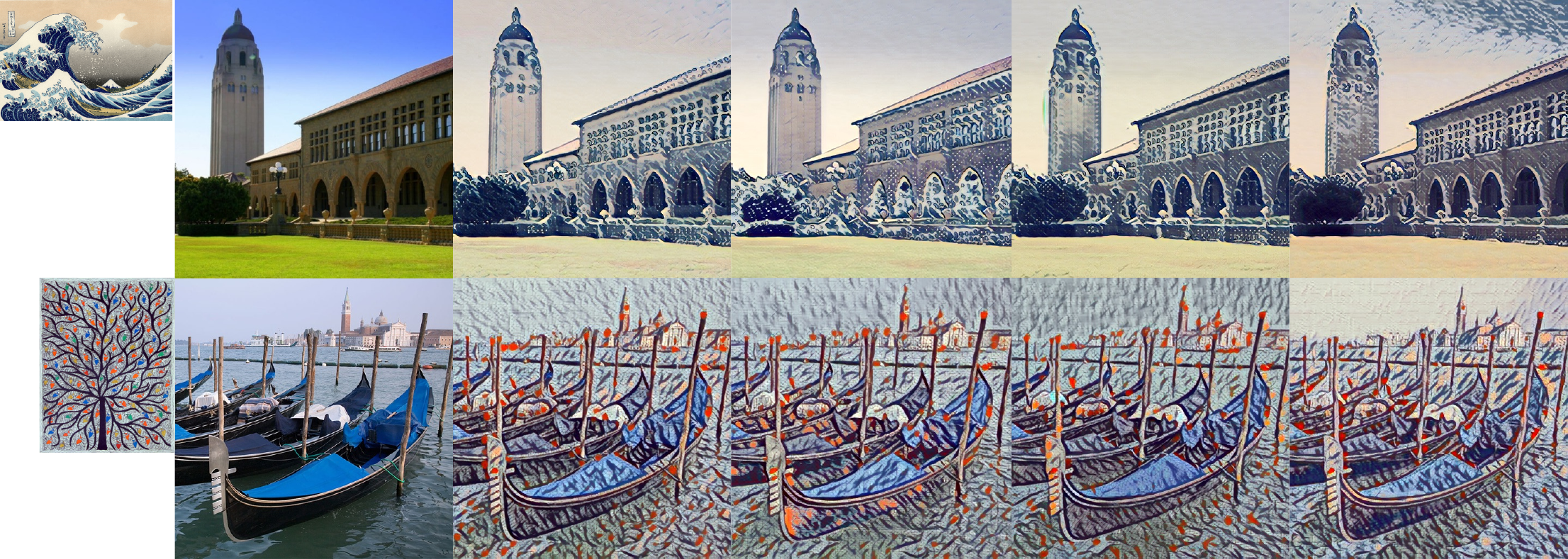}
\end{center}
   \caption{The stylized results of StyleRemix with different number of style set $S$, \ie 5, 10, 15, 50.}
\label{fig:compare_styleNumbers} 
\end{figure}

Style scalability is critical to SEL methods. However, it is difficult to measure since the maximum capabilities of a single model is highly related to the particular set of styles.
For example, a SEL method is likely to have good style scalability if all styles come from one domain instead of different sources.
To evaluate the capacity of StyleRemix, we experiment with the same diverse style set $S_{50}$ as Section \ref{sec:MST}
and study how varying the number of styles affects the stylization.
We train StyleRemix with subsets of $S_{50}$, \ie $S_{5} \subset S_{10} \subset S_{15}$, with the same parameters.
According to Fig. \ref{fig:compare_styleNumbers}, as the number of styles increases, the result shows visually consistent stylization effect though some details of style may fade progressively. This phenomena indicates that StyleRemix can handle a certain number of diverse styles without significant visual degradation. 



\section{Conclusion}
In this paper, we have proposed a style embedding learning approach, StyleRemix, for multi-style transfer task. 
Combining interpretability with efficiency, our method effectively perform multi-style transfer and produce  visually appealing stylization. In addition, the proposed method can be extended style visualization task. 
More importantly, StyleRemix provides several ways to remix styles in the style embedding space. Experimental results demonstrate the effectiveness of StyleRemix on many multi-style transfer tasks. As future direction, one may extend StyleRemix to handle arbitrary style transfer.




\newpage

{\small
\bibliographystyle{ieee}
\bibliography{egbib}

\begin{thebibliography}{10}\itemsep=-1pt

\bibitem{chen2017stylebank}
D.~Chen, L.~Yuan, J.~Liao, N.~Yu, and G.~Hua.
\newblock Stylebank: An explicit representation for neural image style
  transfer.
\newblock In {\em Proceedings of the IEEE Conference on Computer Vision and
  Pattern Recognition (CVPR)}, volume~1, page~4, 2017.

\bibitem{chen2016fast}
T.~Q. Chen and M.~Schmidt.
\newblock Fast patch-based style transfer of arbitrary style.
\newblock {\em Advances in Neural Information Processing Systems (NPIS),
  Constructive Machine Learning Workshop}, 2016.

\bibitem{dumoulin2017learned}
V.~Dumoulin, J.~Shlens, and M.~Kudlur.
\newblock A learned representation for artistic style.
\newblock {\em Proceedings of International Conference on Learning
  Representations (ICLR)}, 2017.

\bibitem{efros1999texture}
A.~A. Efros and T.~K. Leung.
\newblock Texture synthesis by non-parametric sampling.
\newblock In {\em IEEE International Conference on Computer Vision (ICCV)},
  page 1033. IEEE, 1999.

\bibitem{gatys2015texture}
L.~Gatys, A.~S. Ecker, and M.~Bethge.
\newblock Texture synthesis using convolutional neural networks.
\newblock In {\em Advances in Neural Information Processing Systems (NIPS)},
  pages 262--270, 2015.

\bibitem{gatys2016image}
L.~A. Gatys, A.~S. Ecker, and M.~Bethge.
\newblock Image style transfer using convolutional neural networks.
\newblock In {\em IEEE Conference on Computer Vision and Pattern Recognition
  (CVPR)}, pages 2414--2423. IEEE, 2016.

\bibitem{ghiasi2017exploring}
G.~Ghiasi, H.~Lee, M.~Kudlur, V.~Dumoulin, and J.~Shlens.
\newblock Exploring the structure of a real-time, arbitrary neural artistic
  stylization network.
\newblock {\em arXiv preprint arXiv:1705.06830}, 2017.

\bibitem{goodfellow2014generative}
I.~Goodfellow, J.~Pouget-Abadie, M.~Mirza, B.~Xu, D.~Warde-Farley, S.~Ozair,
  A.~Courville, and Y.~Bengio.
\newblock Generative adversarial nets.
\newblock In {\em Advances in neural information processing systems (NIPS)},
  pages 2672--2680, 2014.

\bibitem{gu2018arbitrary}
S.~Gu, C.~Chen, J.~Liao, and L.~Yuan.
\newblock Arbitrary style transfer with deep feature reshuffle.
\newblock In {\em Proceedings of the IEEE Conference on Computer Vision and
  Pattern Recognition (CVPR)}, pages 8222--8231, 2018.

\bibitem{hertzmann2001image}
A.~Hertzmann, C.~E. Jacobs, N.~Oliver, B.~Curless, and D.~H. Salesin.
\newblock Image analogies.
\newblock In {\em Proceedings of the 28th annual conference on Computer
  graphics and interactive techniques}, pages 327--340. ACM, 2001.

\bibitem{hinton2006reducing}
G.~E. Hinton and R.~R. Salakhutdinov.
\newblock Reducing the dimensionality of data with neural networks.
\newblock {\em science}, 313(5786):504--507, 2006.

\bibitem{huang2017arbitrary}
X.~Huang and S.~J. Belongie.
\newblock Arbitrary style transfer in real-time with adaptive instance
  normalization.
\newblock In {\em IEEE International Conference on Computer Vision (ICCV)},
  pages 1510--1519, 2017.

\bibitem{isola2017image}
P.~Isola, J.-Y. Zhu, T.~Zhou, and A.~A. Efros.
\newblock Image-to-image translation with conditional adversarial networks.
\newblock {\em Proceedings of the IEEE Conference on Computer Vision and
  Pattern Recognition (CVPR)}, 2017.

\bibitem{johnson2016perceptual}
J.~Johnson, A.~Alahi, and L.~Fei-Fei.
\newblock Perceptual losses for real-time style transfer and super-resolution.
\newblock In {\em European Conference on Computer Vision (ECCV)}, pages
  694--711. Springer, 2016.

\bibitem{karacan2016learning}
L.~Karacan, Z.~Akata, A.~Erdem, and E.~Erdem.
\newblock Learning to generate images of outdoor scenes from attributes and
  semantic layouts.
\newblock {\em arXiv preprint arXiv:1612.00215}, 2016.

\bibitem{karayev2013recognizing}
S.~Karayev, M.~Trentacoste, H.~Han, A.~Agarwala, T.~Darrell, A.~Hertzmann, and
  H.~Winnemoeller.
\newblock Recognizing image style.
\newblock {\em arXiv preprint arXiv:1311.3715}, 2013.

\bibitem{kim2017learning}
T.~Kim, M.~Cha, H.~Kim, J.~K. Lee, and J.~Kim.
\newblock Learning to discover cross-domain relations with generative
  adversarial networks.
\newblock {\em arXiv preprint arXiv:1703.05192}, 2017.

\bibitem{kingma2014adam}
D.~P. Kingma and J.~Ba.
\newblock Adam: A method for stochastic optimization.
\newblock {\em Proceedings of the International Conference on Learning
  Representations (ICLR)}, 2014.

\bibitem{kingma2013auto}
D.~P. Kingma and M.~Welling.
\newblock Auto-encoding variational bayes.
\newblock {\em Proceedings of the International Conference on Learning
  Representations (ICLR)}, 2013.

\bibitem{li2018learning}
X.~Li, S.~Liu, J.~Kautz, and M.-H. Yang.
\newblock Learning linear transformations for fast arbitrary style transfer.
\newblock {\em arXiv preprint arXiv:1808.04537}, 2018.

\bibitem{li2017diversified}
Y.~Li, C.~Fang, J.~Yang, Z.~Wang, X.~Lu, and M.-H. Yang.
\newblock Diversified texture synthesis with feed-forward networks.
\newblock In {\em Proceedings of the IEEE Conference on Computer Vision and
  Pattern Recognition (CVPR)}, 2017.

\bibitem{liao2017visual}
J.~Liao, Y.~Yao, L.~Yuan, G.~Hua, and S.~B. Kang.
\newblock Visual attribute transfer through deep image analogy.
\newblock {\em ACM Transactions on Graphics (TOG)}, 2017.

\bibitem{lin2014microsoft}
T.-Y. Lin, M.~Maire, S.~Belongie, J.~Hays, P.~Perona, D.~Ramanan,
  P.~Doll{\'a}r, and C.~L. Zitnick.
\newblock Microsoft coco: Common objects in context.
\newblock In {\em European Conference on Computer Vision (ECCV)}, pages
  740--755. Springer, 2014.

\bibitem{liu2017unsupervised}
M.-Y. Liu, T.~Breuel, and J.~Kautz.
\newblock Unsupervised image-to-image translation networks.
\newblock In {\em Advances in Neural Information Processing Systems (NIPS)},
  pages 700--708, 2017.

\bibitem{liu2016coupled}
M.-Y. Liu and O.~Tuzel.
\newblock Coupled generative adversarial networks.
\newblock In {\em Advances in neural information processing systems (NIPS)},
  pages 469--477, 2016.

\bibitem{long2015fully}
J.~Long, E.~Shelhamer, and T.~Darrell.
\newblock Fully convolutional networks for semantic segmentation.
\newblock In {\em Proceedings of the IEEE Conference on Computer Vision and
  Pattern Recognition (CVPR)}, pages 3431--3440, 2015.

\bibitem{maaten2008visualizing}
L.~v.~d. Maaten and G.~Hinton.
\newblock Visualizing data using t-sne.
\newblock {\em Journal of machine learning research (JMLR)}, 9(Nov):2579--2605,
  2008.

\bibitem{paszke2017automatic}
A.~Paszke, S.~Gross, S.~Chintala, G.~Chanan, E.~Yang, Z.~DeVito, Z.~Lin,
  A.~Desmaison, L.~Antiga, and A.~Lerer.
\newblock Automatic differentiation in pytorch.
\newblock 2017.

\bibitem{russakovsky2015imagenet}
O.~Russakovsky, J.~Deng, H.~Su, J.~Krause, S.~Satheesh, S.~Ma, Z.~Huang,
  A.~Karpathy, A.~Khosla, M.~Bernstein, et~al.
\newblock Imagenet large scale visual recognition challenge.
\newblock {\em International Journal of Computer Vision (IJCV)},
  115(3):211--252, 2015.

\bibitem{sanakoyeu2018style}
A.~Sanakoyeu, D.~Kotovenko, S.~Lang, and B.~Ommer.
\newblock A style-aware content loss for real-time hd style transfer.
\newblock {\em European Conference on Computer Vision (ECCV)}, 2018.

\bibitem{sangkloy2017scribbler}
P.~Sangkloy, J.~Lu, C.~Fang, F.~Yu, and J.~Hays.
\newblock Scribbler: Controlling deep image synthesis with sketch and color.
\newblock In {\em IEEE Conference on Computer Vision and Pattern Recognition
  (CVPR)}, volume~2, 2017.

\bibitem{shen2018neural}
F.~Shen, S.~Yan, and G.~Zeng.
\newblock Neural style transfer via meta networks.
\newblock In {\em Proceedings of the IEEE Conference on Computer Vision and
  Pattern Recognition (CVPR)}, pages 8061--8069, 2018.

\bibitem{sheng2018avatar}
L.~Sheng, Z.~Lin, J.~Shao, and X.~Wang.
\newblock Avatar-net: Multi-scale zero-shot style transfer by feature
  decoration.
\newblock In {\em Proceedings of the IEEE Conference on Computer Vision and
  Pattern Recognition (CVPR)}, pages 8242--8250, 2018.

\bibitem{simonyan2014very}
K.~Simonyan and A.~Zisserman.
\newblock Very deep convolutional networks for large-scale image recognition.
\newblock {\em arXiv preprint arXiv:1409.1556}, 2014.

\bibitem{ulyanov2016texture}
D.~Ulyanov, V.~Lebedev, A.~Vedaldi, and V.~S. Lempitsky.
\newblock Texture networks: Feed-forward synthesis of textures and stylized
  images.
\newblock In {\em International Conference on Machine Learning (ICML)}, pages
  1349--1357, 2016.

\bibitem{ulyanov2017improved}
D.~Ulyanov, A.~Vedaldi, and V.~S. Lempitsky.
\newblock Improved texture networks: Maximizing quality and diversity in
  feed-forward stylization and texture synthesis.
\newblock In {\em IEEE Conference on Computer Vision and Pattern Recognition
  (CVPR)}, volume~1, page~3, 2017.

\bibitem{wang2017zm}
H.~Wang, X.~Liang, H.~Zhang, D.-Y. Yeung, and E.~P. Xing.
\newblock Zm-net: Real-time zero-shot image manipulation network.
\newblock {\em arXiv preprint arXiv:1703.07255}, 2017.

\bibitem{wang2017multimodal}
X.~Wang, G.~Oxholm, D.~Zhang, and Y.-F. Wang.
\newblock Multimodal transfer: A hierarchical deep convolutional neural network
  for fast artistic style transfer.
\newblock In {\em Proceedings of the IEEE Conference on Computer Vision and
  Pattern Recognition (CVPR)}, volume~2, page~7, 2017.

\bibitem{wei2000fast}
L.-Y. Wei and M.~Levoy.
\newblock Fast texture synthesis using tree-structured vector quantization.
\newblock In {\em Proceedings of the 27th annual conference on Computer
  Graphics and Interactive Techniques}, pages 479--488. ACM
  Press/Addison-Wesley Publishing Co., 2000.

\bibitem{yanai2017conditional}
K.~Yanai and R.~Tanno.
\newblock Conditional fast style transfer network.
\newblock In {\em Proceedings of the 2017 ACM on International Conference on
  Multimedia Retrieval (ICMR)}, pages 434--437. ACM, 2017.

\bibitem{zhang2017multi}
H.~Zhang and K.~Dana.
\newblock Multi-style generative network for real-time transfer.
\newblock {\em European Conference on Computer Vision (ECCV) Workshops}, 2018.

\bibitem{zhu2017unpaired}
J.-Y. Zhu, T.~Park, P.~Isola, and A.~A. Efros.
\newblock Unpaired image-to-image translation using cycle-consistent
  adversarial networks.
\newblock {\em Proceedings of the IEEE International Conference on Computer
  Vision (CVPR)}, 2017.

\end{thebibliography}
}

\end{document}